%% file: main.tex
\documentclass[smallextended, final]{svjour3}
\usepackage[utf8]{inputenc}
\usepackage{appendix}

\usepackage{graphicx}
\usepackage{amssymb}
\usepackage{soul}

\usepackage{floatrow}
\usepackage{amsfonts}
\usepackage{mathtools}
\usepackage{graphicx}
\usepackage[tight,footnotesize]{subfigure}
\usepackage{multirow}
\usepackage{array}
\usepackage{multicol}
\usepackage{url}

\newfloatcommand{capbtabbox}{table}[][\FBwidth]
\usepackage{amssymb}
\usepackage{amsthm}
\usepackage{xspace}

\usepackage[table]{xcolor}
\xdefinecolor{gray95}{gray}{0.65}
\xdefinecolor{gray25}{gray}{0.8}

 \usepackage{algpseudocode}
 \usepackage{mathtools}
\usepackage{caption}
 \usepackage{rotating}
 \usepackage{pdflscape}
 \usepackage[linesnumbered,ruled]{algorithm2e}
 \SetKwRepeat{Do}{do}{while}%

 \usepackage{todonotes}

\title{CMSA Algorithm for Solving the Prioritized Pairwise Test Data Generation Problem in Software Product Lines}

\begin{document}

\author{Javier Ferrer
\and Francisco Chicano
\and Jos\'e Antonio Ortega Toro
}

\institute{ Javier Ferrer and Francisco Chicano \at
Universidad de M\'alaga, España \\
 \email{ferrer, chicano@lcc.uma.es}
\and Jos\'e Antonio Ortega-Toro \at
CERN, Suiza \\
 \email{ortega.toro@cern.ch}
}

\date{Received: date / Accepted: date}

\maketitle

\begin{abstract}
In Software Product Lines (SPLs) it may be difficult or even impossible to test all the products of the family because of the large number of valid feature combinations that may exist \cite{Ferrer2017}. Thus, we want to find a minimal subset of the product family that allows us to test all these possible combinations (pairwise). Furthermore, when testing a single product is a great effort, it is desirable to first test products composed of a set of priority features. This problem is called Prioritized Pairwise Test Data Generation Problem. 

State-of-the-art algorithms based on Integer Linear Programming for this problema are faster enough for small and medium instances. However, there exists some real instances that are too large to be computed with these algorithms in a reasonable time because of the exponential growth of the number of candidate solutions. Also, these heuristics not always lead us to the best solutions. In this work we propose a new approach based on a hybrid metaheuristic algorithm called \textit{Construct, Merge, Solve \& Adapt}. We compare this matheuristic with four algorithms: a Hybrid algorithm based on Integer Linear Programming ((HILP), a Hybrid algorithm based on Integer Nonlinear Programming (HINLP), the Parallel Prioritized Genetic Solver (PPGS), and a greedy algorithm called prioritized-ICPL. The analysis reveals that CMSA results in statistically significantly better quality solutions in most instances and for most levels of weighted coverage, although it requires more execution time.

\end{abstract}



\section{Introduction}
\input{Chapters/Introduction}

\section{Background}
\label{sec:background}
\input{Chapters/Background}

\section{Applying CMSA to prioritized SPL}
\input{Chapters/DesignAndImplementation}

\section{Experimental Setup}
\input{Chapters/Results}

\section{Conclusions}
\input{Chapters/Conclusions}

\section*{Acknowledgments}

This research was partially funded by Ministerio de Econom\'ia, Industria y Competitividad, Gobierno de Espa\~na, and European Regional Development Fund grant numbers TIN2016-81766-REDT (\url{http://cirti.es}), and TIN2017-88213-R (\url{http://6city.lcc.uma.es}), and by Universidad de M\'alaga under the EXHAURO (https://exhauro.uma.es) project.



\input{main.bbl}


\appendix

\section*{Appendix}

In this appendix we present the detailed results in Table~\ref{tab:big}. This table shows the mean number of products required to reach all percentages considered of weighted coverage for all the feature models and all the algorithms.

\begin{table}[h!]
\scriptsize
\setlength\tabcolsep{1.5pt}
\centering
\caption{Mean number of products needed to achieve all percentages of weighted coverage for all feature models. A value is highlighted when it is strictly smaller than the others for that particular level of coverage and feature model.}
\label{tab:big}
\begin{tabular}{l|l|l|l|l|l|l|l|l|l|l|l|l|r}
\hline
\multirow{1}{*}{Feature Models} & Algor. & 50\% & 75\%  & 80\% &85\% & 90\% & 95\% & 96\% &97\% &98\% &99\% &100\% & Time (ms)  \\ \hline

                                            & CMSA  & 2.00 & 3.00 & 3.00 & 4.00 & 4.00 & \cellcolor[HTML]{C0C0C0}5.00 & \cellcolor[HTML]{C0C0C0}5.00 & 6.00 & \cellcolor[HTML]{C0C0C0}6.00  & \cellcolor[HTML]{C0C0C0}6.00  & \cellcolor[HTML]{C0C0C0}6.00  & 9522.80     \\ \cline{2-14} 
                                            & HILP  & 2.00 & 3.00 & 3.00 & 4.00 & 4.00 & 6.00 & 6.00 & 6.00 & 7.00  & 7.00  & 7.00                          & 9695.00     \\ \cline{2-14} 
                                            & HINLP & 2.00 & 3.00 & 3.00 & 4.00 & 4.00 & 6.00 & 6.00 & 6.00 & 7.00  & 7.00  & 7.00                          & 449.00      \\ \cline{2-14} 
                                            & pICPL & 2.00 & 3.00 & 3.00 & 4.00 & 5.00 & 6.00 & 7.00 & 7.00 & 7.00  & 8.00  & 8.00                          & 741.60      \\ \cline{2-14} 
\multirow{-5}{*}{Apache}                & PPGS  & 2.00 & 3.00 & 3.00 & 4.00 & 4.00 & 6.00 & 6.00 & 6.00 & 7.00  & 7.00  & 7.00                          & 10394.03    \\ \hline
                                            & CMSA  & 2.00 & 3.00 & 4.00 & \cellcolor[HTML]{C0C0C0}4.00 & \cellcolor[HTML]{C0C0C0}4.00 & \cellcolor[HTML]{C0C0C0}6.00 & \cellcolor[HTML]{C0C0C0}6.00 & \cellcolor[HTML]{C0C0C0}6.00 & \cellcolor[HTML]{C0C0C0}6.00  & \cellcolor[HTML]{C0C0C0}6.07  & \cellcolor[HTML]{C0C0C0}6.07  & 9537.20     \\ \cline{2-14} 
                                            & HILP  & 2.00 & 4.00 & 4.00 & 5.00 & 6.00 & 7.00 & 7.00 & 7.00 & 8.00  & 8.00  & 8.00                          & 19015.00    \\ \cline{2-14} 
                                            & HINLP & 2.00 & 4.00 & 4.00 & 5.00 & 5.00 & 7.00 & 7.00 & 7.00 & 7.00  & 8.00  & 8.00                          & 440.00      \\ \cline{2-14} 
                                            & pICPL & 2.00 & 4.00 & 5.00 & 6.00 & 7.00 & 8.00 & 8.00 & 8.00 & 8.00  & 9.00  & 9.00                          & 687.07      \\ \cline{2-14} 
\multirow{-5}{*}{BDBFootprint}   & PPGS  & 2.00 & 4.00 & 4.00 & 5.00 & 5.97 & 6.97 & 6.97 & 6.97 & 7.97  & 8.00  & 8.17                          & 11213.53    \\ \hline
                                            & CMSA  & 2.00 & 3.00 & 3.00 & 4.00 & 4.00 & \cellcolor[HTML]{C0C0C0}6.00 & \cellcolor[HTML]{C0C0C0}6.73 & \cellcolor[HTML]{C0C0C0}7.27 & \cellcolor[HTML]{C0C0C0}8.20  & \cellcolor[HTML]{C0C0C0}10.00 & \cellcolor[HTML]{C0C0C0}20.03 & 148818.63   \\ \cline{2-14} 
                                            & HILP  & 2.00 & 3.00 & 3.00 & 4.00 & 4.00 & 6.00 & 7.00 & 8.00 & 9.00  & 11.00 & 21.00                         & 21404.00    \\ \cline{2-14} 
                                            & HINLP & 2.00 & 3.00 & 3.00 & 4.00 & 4.00 & 6.00 & 7.00 & 8.00 & 9.00  & 11.00 & 21.00                         & 1010.00     \\ \cline{2-14} 
                                            & pICPL & 2.00 & 3.00 & 3.00 & 4.00 & 6.00 & 7.00 & 8.00 & 8.00 & 10.00 & 11.00 & 21.00                         & 9911.80     \\ \cline{2-14} 
\multirow{-5}{*}{BDBMemory}      & PPGS  & 2.00 & 3.00 & 3.00 & 4.00 & 4.73 & 6.87 & 7.80 & 8.77 & 9.97  & 11.90 & 23.33                         & 117607.53   \\ \hline
                                            & CMSA  & 1.00 & 2.00 & 2.00 & 3.00 & 3.00 & 4.00 & 4.00 & 5.00 & \cellcolor[HTML]{C0C0C0}5.00  & \cellcolor[HTML]{C0C0C0}6.00  & \cellcolor[HTML]{C0C0C0}9.00  & 56589.83    \\ \cline{2-14} 
                                            & HILP  & 1.00 & 2.00 & 2.00 & 3.00 & 3.00 & 4.00 & 4.00 & 5.00 & 6.00  & 7.00  & 10.00                         & 14993.00    \\ \cline{2-14} 
                                            & HINLP & 1.00 & 2.00 & 2.00 & 3.00 & 3.00 & 4.00 & 4.00 & 5.00 & 6.00  & 7.00  & 10.00                         & 577.00      \\ \cline{2-14} 
                                            & pICPL & 1.00 & 2.00 & 3.00 & 3.00 & 4.00 & 6.00 & 6.00 & 6.00 & 6.00  & 7.00  & 12.00                         & 6008.37     \\ \cline{2-14} 
\multirow{-5}{*}{BDBPerformance} & PPGS  & 1.00 & 2.00 & 2.00 & 3.00 & 3.00 & 4.00 & 4.83 & 5.00 & 5.93  & 7.00  & 10.60                         & 47361.73    \\ \hline
                                            & CMSA  & 2.00 & 2.53 & 3.00 & \cellcolor[HTML]{C0C0C0}3.00 & 4.00 & \cellcolor[HTML]{C0C0C0}5.00 & \cellcolor[HTML]{C0C0C0}5.00 & 6.00 & \cellcolor[HTML]{C0C0C0}6.00  & \cellcolor[HTML]{C0C0C0}6.90  & 8.00  & 22736.93    \\ \cline{2-14} 
                                            & HILP  & 2.00 & 3.00 & 3.00 & 4.00 & 4.00 & 6.00 & 6.00 & 6.00 & 7.00  & 8.00  & 9.00                          & 13639.00    \\ \cline{2-14} 
                                            & HINLP & 2.00 & 3.00 & 3.00 & 4.00 & 4.00 & 6.00 & 6.00 & 6.00 & 7.00  & 8.00  & 9.00                          & 451.00      \\ \cline{2-14} 
                                            & pICPL & 2.00 & 3.00 & 3.00 & 4.00 & 4.00 & 6.00 & 6.00 & 6.00 & 7.00  & 7.00  & 8.00  & 648.27      \\ \cline{2-14} 
\multirow{-5}{*}{Curl}                  & PPGS  & 2.00 & 3.00 & 3.00 & 3.97 & 4.03 & 5.83 & 6.00 & 6.50 & 7.37  & 8.07  & 9.63                          & 17454.57    \\ \hline
                                            & CMSA  & 1.00 & 2.00 & 2.00 & 2.00 & 3.00 & 4.00 & 4.00 & 5.00 & \cellcolor[HTML]{C0C0C0}5.00  & \cellcolor[HTML]{C0C0C0}7.00  & \cellcolor[HTML]{C0C0C0}11.10 & 69734.97    \\ \cline{2-14} 
                                            & HILP  & 1.00 & 2.00 & 2.00 & 2.00 & 3.00 & 4.00 & 5.00 & 5.00 & 6.00  & 8.00  & 13.00                         & 12894.00    \\ \cline{2-14} 
                                            & HINLP & 1.00 & 2.00 & 2.00 & 2.00 & 3.00 & 4.00 & 5.00 & 5.00 & 6.00  & 8.00  & 13.00                         & 625.00      \\ \cline{2-14} 
                                            & pICPL & 1.00 & 2.00 & 2.00 & 3.00 & 3.00 & 4.00 & 4.00 & 5.00 & 7.00  & 11.00 & 14.00                         & 6865.53     \\ \cline{2-14} 
\multirow{-5}{*}{LinkedList}            & PPGS  & 1.00 & 2.00 & 2.00 & 2.00 & 3.00 & 4.23 & 5.00 & 5.00 & 6.13  & 7.73  & 13.37                         & 60684.57    \\ \hline
                                            & CMSA  & 2.00 & 4.00 & 4.00 & 5.00 & \cellcolor[HTML]{C0C0C0}5.70 & 7.00 & \cellcolor[HTML]{C0C0C0}7.00 & \cellcolor[HTML]{C0C0C0}7.90 & 8.03  & 9.00  & 11.00                         & 137948.13   \\ \cline{2-14} 
                                            & HILP  & 2.00 & 4.00 & 4.00 & 5.00 & 6.00 & 7.00 & 8.00 & 8.00 & 9.00  & 10.00 & 11.00                         & 29396.00    \\ \cline{2-14} 
                                            & HINLP & 2.00 & 4.00 & 4.00 & 5.00 & 6.00 & 7.00 & 8.00 & 8.00 & 9.00  & 10.00 & 11.00                         & 6813.00     \\ \cline{2-14} 
                                            & pICPL & 2.00 & 4.00 & 5.00 & 5.00 & 6.00 & 8.00 & 8.00 & 8.00 & \cellcolor[HTML]{C0C0C0}8.00  & 9.00  & \cellcolor[HTML]{C0C0C0}10.00 & 3539.37     \\ \cline{2-14} 
\multirow{-5}{*}{Linux}                 & PPGS  & 2.00 & 4.00 & 4.00 & 5.00 & 6.00 & 7.00 & 7.67 & 8.00 & 8.37  & 9.40  & 11.10                         & 49385.43    \\ \hline
                                            & CMSA  & 2.00 & 3.00 & \cellcolor[HTML]{C0C0C0}3.00 & 4.00 & \cellcolor[HTML]{C0C0C0}4.00 & \cellcolor[HTML]{C0C0C0}5.00 & 6.00 & \cellcolor[HTML]{C0C0C0}6.00 & \cellcolor[HTML]{C0C0C0}6.00  & \cellcolor[HTML]{C0C0C0}6.00  & \cellcolor[HTML]{C0C0C0}6.90  & 15747.90    \\ \cline{2-14} 
                                            & HILP  & 2.00 & 3.00 & 4.00 & 4.00 & 5.00 & 6.00 & 6.00 & 7.00 & 7.00  & 8.00  & 10.00                         & 19164.00    \\ \cline{2-14} 
                                            & HINLP & 2.00 & 3.00 & 4.00 & 4.00 & 5.00 & 6.00 & 6.00 & 7.00 & 7.00  & 8.00  & 10.00                         & 472.00      \\ \cline{2-14} 
                                            & pICPL & 2.00 & 3.00 & 4.00 & 4.00 & 5.00 & 6.00 & 7.00 & 7.00 & 7.00  & 8.00  & 8.00                          & 526.50      \\ \cline{2-14} 
\multirow{-5}{*}{LLVM}                  & PPGS  & 2.00 & 3.00 & 3.03 & 4.00 & 5.00 & 6.00 & 6.00 & 6.07 & 7.00  & 8.00  & 8.17                          & 12805.90    \\ \hline
                                            & CMSA  & 1.00 & 2.00 & 2.00 & 3.00 & 3.00 & \cellcolor[HTML]{C0C0C0}4.00 & \cellcolor[HTML]{C0C0C0}4.00 & 5.00 & 5.00  & \cellcolor[HTML]{C0C0C0}5.00  & \cellcolor[HTML]{C0C0C0}6.00  & 8524.03     \\ \cline{2-14} 
                                            & HILP  & 1.00 & 2.00 & 2.00 & 3.00 & 3.00 & 5.00 & 5.00 & 5.00 & 5.00  & 6.00  & 7.00                          & 6137.00     \\ \cline{2-14} 
                                            & HINLP & 1.00 & 2.00 & 2.00 & 3.00 & 3.00 & 5.00 & 5.00 & 5.00 & 5.00  & 6.00  & 7.00                          & 368.00      \\ \cline{2-14} 
                                            & pICPL & 1.00 & 2.00 & 3.00 & 3.00 & 3.00 & 5.00 & 5.00 & 6.00 & 7.00  & 8.00  & 8.00                          & 501.13      \\ \cline{2-14} 
\multirow{-5}{*}{PKJab}                 & PPGS  & 1.00 & 2.00 & 2.00 & 3.00 & 3.07 & 4.53 & 5.00 & 5.00 & 5.17  & 6.00  & 7.00                          & 11439.47    \\ \hline
                                            & CMSA  & 2.00 & 3.00 & 3.00 & 3.00 & 4.00 & 5.00 & 5.00 & 5.00 & \cellcolor[HTML]{C0C0C0}5.00  & 6.00  & 6.00                          & 4755.83     \\ \cline{2-14} 
                                            & HILP  & 2.00 & 3.00 & 3.00 & 3.00 & 4.00 & 5.00 & 5.00 & 6.00 & 6.00  & 6.00  & 6.00                          & 8086.00     \\ \cline{2-14} 
                                            & HINLP & 2.00 & 3.00 & 3.00 & 3.00 & 4.00 & 5.00 & 5.00 & 6.00 & 6.00  & 6.00  & 6.00                          & 325.00      \\ \cline{2-14} 
                                            & pICPL & 2.00 & 3.00 & 3.00 & 3.00 & 4.00 & 5.00 & 5.00 & 5.00 & 6.00  & 6.00  & 6.00                          & 238.37      \\ \cline{2-14} 
\multirow{-5}{*}{Prevayler}             & PPGS  & 2.00 & 3.00 & 3.00 & 3.00 & 4.00 & 5.00 & 5.00 & 5.60 & 6.00  & 6.00  & 6.00                          & 8091.17     \\ \hline
                                            & CMSA  & 1.00 & \cellcolor[HTML]{C0C0C0}2.97 & 3.00 & 3.00 & 4.00 & 5.00 & 5.00 & \cellcolor[HTML]{C0C0C0}5.20 & \cellcolor[HTML]{C0C0C0}6.00  & \cellcolor[HTML]{C0C0C0}6.93  & \cellcolor[HTML]{C0C0C0}10.00 & 155619.87   \\ \cline{2-14} 
                                            & HILP  & 1.00 & 3.00 & 3.00 & 3.00 & 4.00 & 5.00 & 5.00 & 6.00 & 7.00  & 8.00  & 14.00                         & 21154.00    \\ \cline{2-14} 
                                            & HINLP & 1.00 & 3.00 & 3.00 & 3.00 & 4.00 & 5.00 & 5.00 & 6.00 & 7.00  & 8.00  & 14.00                         & 951.00      \\ \cline{2-14} 
                                            & pICPL & 1.00 & 3.00 & 4.00 & 5.00 & 6.00 & 8.00 & 9.00 & 9.00 & 10.00 & 11.00 & 17.00                         & 74181.93    \\ \cline{2-14} 
\multirow{-5}{*}{SensorNetwork}         & PPGS  & 1.00 & 3.00 & 3.00 & 3.00 & 4.00 & 5.03 & 5.47 & 6.00 & 6.97  & 7.87  & 13.97                         & 71971.50    \\ \hline
                                            & CMSA  & 1.00 & 2.00 & 2.00 & 3.00 & 4.00 & \cellcolor[HTML]{C0C0C0}5.00 & 6.00 & \cellcolor[HTML]{C0C0C0}6.37 & \cellcolor[HTML]{C0C0C0}7.57  & \cellcolor[HTML]{C0C0C0}9.30  & \cellcolor[HTML]{C0C0C0}23.87 & 1416002.00  \\ \cline{2-14} 
                                            & HILP  & 1.00 & 2.00 & 2.00 & 3.00 & 4.00 & 6.00 & 6.00 & 7.00 & 8.00  & 10.00 & 28.00                         & 47816.00    \\ \cline{2-14} 
                                            & HINLP & 1.00 & 2.00 & 2.00 & 3.00 & 4.00 & 6.00 & 6.00 & 7.00 & 8.00  & 10.00 & 27.00                         & 6450.00     \\ \cline{2-14} 
                                            & pICPL & 1.00 & 3.00 & 4.00 & 4.00 & 5.00 & 8.00 & 8.00 & 9.00 & 11.00 & 14.00 & 28.00                         & 41312.30    \\ \cline{2-14} 
\multirow{-5}{*}{SQLiteMemory}          & PPGS  & 1.03 & 2.17 & 2.90 & 3.23 & 4.07 & 6.03 & 6.97 & 7.93 & 9.23  & 11.70 & 31.53                         & 903118.97   \\ \hline
                                            & CMSA  & 1.00 & 1.00 & 1.00 & 2.00 & 2.00 & 3.00 & 3.00 & 3.00 & 3.00  & 4.00  & 12.00 & 483297.43   \\ \cline{2-14} 
                                            & HILP  & 1.00 & 1.00 & 1.00 & 2.00 & 2.00 & 3.00 & 3.00 & 3.00 & 3.00  & 4.00  & 12.00 & 31558.00    \\ \cline{2-14} 
                                            & HINLP & 1.00 & 1.00 & 1.00 & 2.00 & 2.00 & 3.00 & 3.00 & 3.00 & 3.00  & 4.00  & 12.00 & 2878.00     \\ \cline{2-14} 
                                            & pICPL & 1.00 & 1.00 & 1.00 & 2.00 & 2.00 & 3.00 & 3.00 & 4.00 & 4.00  & 6.00  & 15.00                         & 241170.97   \\ \cline{2-14} 
\multirow{-5}{*}{Violet}                & PPGS  & 1.00 & 1.00 & 1.00 & 2.00 & 2.00 & \cellcolor[HTML]{C0C0C0}2.93 & 3.00 & 3.07 & 3.30  & 4.53  & 12.83                         & 31376054.20 \\ \hline
                                            & CMSA  & 2.00 & 2.00 & 3.00 & 3.00 & 4.00 & 5.00 & \cellcolor[HTML]{C0C0C0}5.97 & 6.00 & \cellcolor[HTML]{C0C0C0}6.63  & \cellcolor[HTML]{C0C0C0}7.40  & \cellcolor[HTML]{C0C0C0}9.87  & 46015.90    \\ \cline{2-14} 
                                            & HILP  & 2.00 & 2.00 & 3.00 & 3.00 & 4.00 & 5.00 & 6.00 & 6.00 & 7.00  & 8.00  & 12.00                         & 16141.00    \\ \cline{2-14} 
                                            & HINLP & 2.00 & 2.00 & 3.00 & 3.00 & 4.00 & 5.00 & 6.00 & 6.00 & 7.00  & 8.00  & 12.00                         & 478.00      \\ \cline{2-14} 
                                            & pICPL & 2.00 & 3.00 & 3.00 & 4.00 & 4.00 & 6.00 & 6.00 & 7.00 & 7.00  & 9.00  & 11.00                         & 1337.43     \\ \cline{2-14} 
\multirow{-5}{*}{Wget}                  & PPGS  & 2.00 & 2.13 & 3.00 & 3.07 & 4.00 & 5.43 & 6.00 & 6.40 & 7.00  & 8.03  & 11.37                         & 31525.37    \\ \hline
                                            & CMSA  & 1.00 & 2.00 & 3.00 & 3.00 & 4.00 & 5.00 & \cellcolor[HTML]{C0C0C0}5.00 & 6.00 & \cellcolor[HTML]{C0C0C0}6.00  & \cellcolor[HTML]{C0C0C0}7.00  & \cellcolor[HTML]{C0C0C0}9.07  & 36472.90    \\ \cline{2-14} 
                                            & HILP  & 1.00 & 2.00 & 3.00 & 3.00 & 4.00 & 5.00 & 6.00 & 6.00 & 7.00  & 8.00  & 12.00                         & 8547.00     \\ \cline{2-14} 
                                            & HINLP & 1.00 & 2.00 & 3.00 & 3.00 & 4.00 & 5.00 & 6.00 & 6.00 & 7.00  & 8.00  & 12.00                         & 479.00      \\ \cline{2-14} 
                                            & pICPL & 1.00 & 2.00 & 3.00 & 3.00 & 4.00 & 5.00 & 6.00 & 7.00 & 7.00  & 9.00  & 13.00                         & 1224.70     \\ \cline{2-14} 
\multirow{-5}{*}{x264}                  & PPGS  & 1.23 & 2.23 & 3.00 & 3.07 & 4.00 & 5.30 & 6.00 & 6.50 & 7.23  & 8.47  & 12.10                         & 37368.53    \\ \hline
                                            & CMSA  & 2.00 & 3.00 & 3.00 & 4.00 & \cellcolor[HTML]{C0C0C0}4.00 & \cellcolor[HTML]{C0C0C0}5.00 & \cellcolor[HTML]{C0C0C0}5.00 & \cellcolor[HTML]{C0C0C0}5.00 & \cellcolor[HTML]{C0C0C0}6.00  & \cellcolor[HTML]{C0C0C0}6.00  & \cellcolor[HTML]{C0C0C0}6.00  & 7097.73     \\ \cline{2-14} 
                                            & HILP  & 2.00 & 3.00 & 3.00 & 4.00 & 5.00 & 6.00 & 6.00 & 7.00 & 7.00  & 7.00  & 7.00                          & 12562.00    \\ \cline{2-14} 
                                            & HINLP & 2.00 & 3.00 & 3.00 & 4.00 & 5.00 & 6.00 & 6.00 & 7.00 & 7.00  & 7.00  & 7.00                          & 355.00      \\ \cline{2-14} 
                                            & pICPL & 2.00 & 3.00 & 3.00 & 4.00 & 5.00 & 6.00 & 6.00 & 6.00 & 7.00  & 7.00  & 7.00                          & 384.50      \\ \cline{2-14} 
\multirow{-5}{*}{ZipMe}                 & PPGS  & 2.00 & 3.00 & 3.00 & 4.00 & 5.00 & 6.00 & 6.00 & 7.00 & 7.00  & 7.00  & 7.03                          & 13035.17    \\ \hline
\end{tabular}
\end{table}

\end{document}

%% file: Chapters/Introduction.tex
Software Product Lines (SPLs) are used to achieve a more efficient software development and management of the variability of software products, reducing the costs and time to market, as well as maintenance costs \cite{pohl2005software}. These product lines carry a great variability within products of the same family of products. This variability is because of the mass customization and it implies a great challenge when we face the task of testing because of the combinatorial explosion in the number of products \cite{ENGSTROM20112}.

Many proposals have arisen having into account these difficulties \cite{Al-Hajjaji2014,Cohen2008}. Some of these are based on \textit{pairwise testing} \cite{perrouin2012pairwise,Oster2010,lopez2013multi}, where each possible combination of two features must be present in at least one product. Some combinations can be more important than others, inducing a priority among configurations or features. In this case, a weight is assigned to each configuration, which can be derived from weights assigned to products. The optimization problem that we want to solve consists in finding a set of products with the minimum cardinal that covers the maximum value of these weighted configurations. Additionally, we want to sort the products in such a way that we first test the products containing higher priority features.

Last state-of-the-art proposals on pairwise testing include hybrid algorithms, mixing heuristics with exact algorithms \cite{Ferrer2017}. This method of testing and its coverage metric are formalized through \textit{feature models}, where products of a family are defined by a unique combination of features \cite{Batory2005}.

The hypothesis for this work is that a hybrid matheuristic approach can improve the performance on large instances of the problem, particularly generating in a probabilistic way new sub-instances of the problem that are combined and later on solved with an exact algorithm in order to find the best solutions to the whole problem.

Our main contribution is the adaptation of a new approach based on a matheuristic named \textit{Construct, Merge, Solve and Adapt} to solve the Prioritized Pairwise Test Data Generation Problem. In order to validate the benefits of our proposal we compare the results with four algorithms: two hybrid algorithms based on integer programming~\cite{Ferrer2017}, one with a linear formulation (HILP) and the other one with a nonlinear formulation (HINLP), the Parallel Prioritized Genetic Solver (PPGS)~\cite{Lopez-Herrejon:2014:PEA:2576768.2598305} and a greedy algorithm called prioritized-ICPL~\cite{DBLP:conf/models/JohansenHFES12}.

The rest of the article is divided into six sections. In the next section we introduce the background required to understand both the problem and the algorithm that we propose. In section~\ref{design} we explain the design and implementation of the CMSA adaptation to the Prioritized Pairwise Test Data Generation Problem. In Section~\ref{experiments} we briefly explain the algorithms in the comparison and the experimental setup. Results are analyzed in Section~\ref{results} and we present our conclusions in Section~\ref{conclusions}.



%% file: Chapters/Background.tex

\subsection{Feature Models}
    Feature models are used in SPLs to define the functionality of a product within a software family as a single combination of features. Models can also represent the constraints that exist between features. A hierarchical tree structure is used to characterize a feature model, where the nodes of the tree are features and the edges represent relationships between these features.

    Figure~\ref{fig:graph} represents the feature model of a classical problem in the evaluation of product-line methodologies, the \textit{Graph Product Line} (GPL) \cite{lopez2001standard}. There exist four types of relationships between features, differentiated graphically in the feature model:
    \begin{itemize}
        \item \textit{Mandatory}: These features are selected when their parent is selected. For example, in the Graph Product Line model, \textit{Driver, Benchmark, GraphType} and \textit{Algorithms} are present in all the products of the family.
        \item \textit{Optional}: They can be or not be selected, like for example, the \textit{Weight} or \textit{Search} features.
        \item \textit{XOR relations}: In these cases only one of the features of the group must be selected when the parent feature is selected. \textit{DFS} and \textit{BFS} features are of this kind in the Graph Product Line model.
        \item \textit{Inclusive-or}: This last kind of relation indicates that at least one of the features of the group must be selected when its parent feature is selected. In the example, when the feature \textit{Algorithms} is selected, at least one of the group $\{$\textit{Num, CC, SCC, Cycle, Shortest, Prim} and \textit{Kruskal}$\}$ must be selected.
    \end{itemize}

There are two other types of constraints over the model, called \textit{Cross-Tree Constraints (CTC)}: \emph{requires} and \emph{excludes}. In the Graph Product Line feature model, we can observe one of these constraints below the tree structure. For example ``Num requires Search'' implies that when the \textit{Num} feature is selected, the \textit{Search} feature must be selected as well. It can easily be seen that these constraints can be formalized using propositional logic.

    \begin{figure}[ht]
        \centering
        \includegraphics[width=\textwidth]{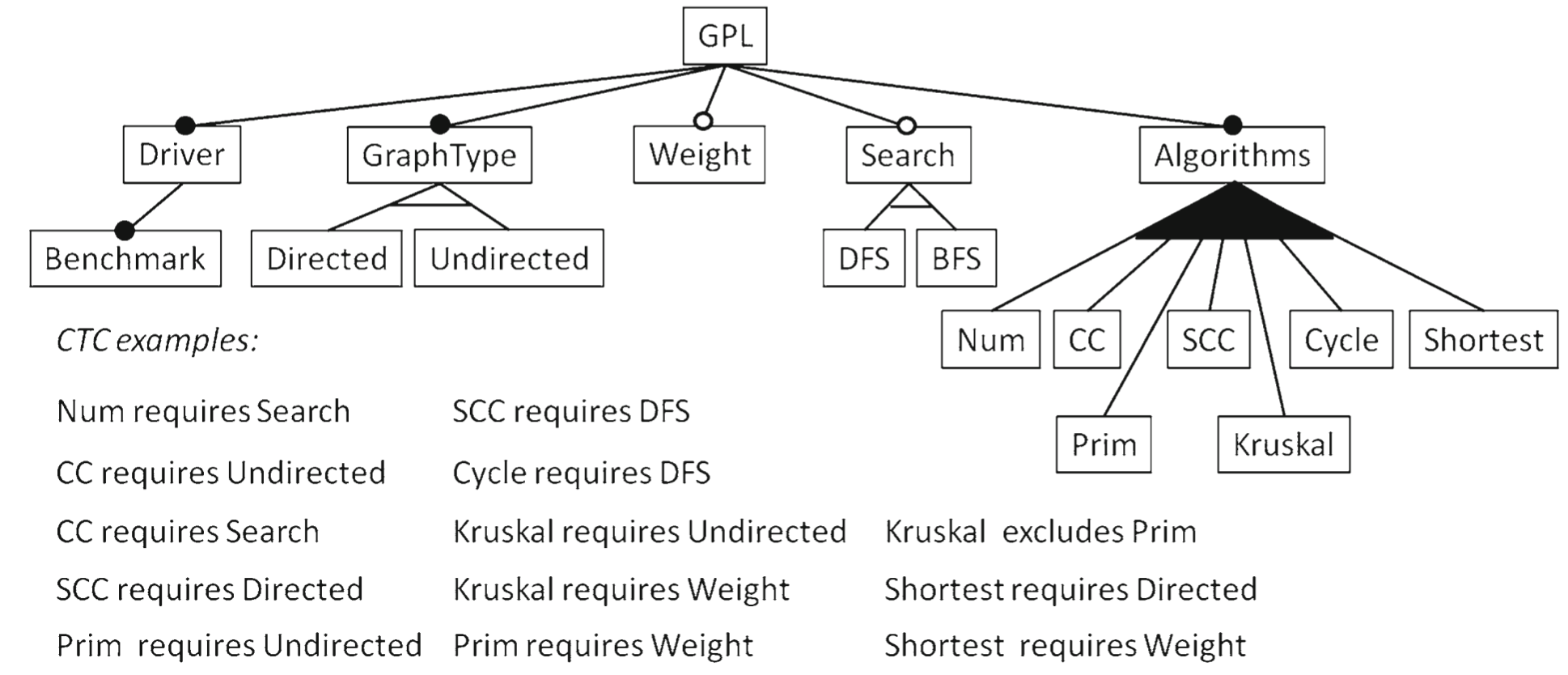}
        \caption[Graph Product Line feature model]{Graph Product Line feature model (from Ferrer et al. \protect\cite{Ferrer2017})}
        \label{fig:graph}
    \end{figure}

\subsection{Problem Formalization: Prioritized Pairwise Test Data Generation}
    \label{sec:problem}
    Now we present the terminology related to \textit{Combinatorial Interaction Testing (CIT)}. This approach builds a set of samples that allow to test different system configurations \cite{Nie:2011:SCT:1883612.1883618}. When we apply this approach to SPL testing, the set of samples is a subset of the products of the family. Next, we introduce these concepts that will lead us to the Prioritized Pairwise Test Data Generation Problem:

    \begin{definition}{\textbf{Feature list}.}
        A feature list $fl$ is the list of all the features in a feature model.
    \end{definition}

    \begin{definition}{\textbf{Product}.}
        A product is represented by a pair $(S, \bar{S})$, where $S$ is a subset of a feature list $FL$, $S \subseteq FL$. Thus, $\bar{S} = FL - S$.
    \end{definition}

    \begin{definition}{\textbf{Valid product}.}
        We say that a product $p$ is valid with respect a feature model $fm$ iff $p.S$ and $p.\bar{S}$ do not violate any constraint described by the feature model. The set of all valid products of a feature model is denoted by $P^{fm}$.
    \end{definition}

    \begin{definition}{\textbf{Pair}.}
        A pair $pr$ is a tuple $(s, \bar{s})$, where $s$ and $\bar{s}$ represent different single features from a feature list $fl$, that is, $pr.s, pr.\bar{s} \in fl\ \land\ pr.s \neq pr.\bar{s}$. A pair is covered by a product $p$ iff $pr.s \in p.S\ \land\ pr.\bar{s} \in p.\bar{S}$. Note that a pair can be presented in the following ways: $(s_1, s_2), (s_1, \bar{s_2}), (\bar{s_1}, s_2), (\bar{s_1}, \bar{s_2})$.
    \end{definition}

    \begin{definition}{\textbf{Valid Pair}.}
        A pair $pr$ is valid within a feature model $fm$ if there exists a product that covers $pr$. The set of all valid pairs in a feature model $fm$ is denoted with $VPR^{fm}$.
    \end{definition}

    \begin{definition}{\textbf{Test suite}.}
        A test suite $ts$ for a feature model $fm$ is a set of valid products of $fm$. A test suite is complete if it covers all the valid pairs in $VPR^{fm}$: $\{\ ts\ |\ \forall pr \in VPR^{fm} \rightarrow \exists p \in ts$ such that $p$ covers $pr$ $\}$
    \end{definition}

    \begin{definition}{\textbf{Prioritized product}.}
        A prioritized product $pp$ is a tuple $(p, w)$, where $p$ represents a valid product in a feature model $fm$ and $w \in \mathbb{R}$ represents its weight.
    \end{definition}

    \begin{definition}{\textbf{Configuration}.}
        A configuration $c$ is a tuple $(pr, w)$ where $pr$ is a valid pair and $w \in \mathbb{R}$ represent its weight. $w$ is computed as follows. Let $PP$ be the set of all prioritized products and $PP_{pr}$ a subset of $PP$, such that $PP_{pr}$ contains all the prioritized products of $PP$ that cover $pr$, that is, $PP_{pr} = \{\ p \in PP\ |\ p $ covers $ pr\ \}$. Then $w = \sum_{p\in PP_{pr}} p.w$
    \end{definition}

    \begin{definition}{\textbf{Covering Array}.}
        A covering array $CA$ for a feature model $fm$ and a set of configurations $C$ is set of valid products $P$ that covers all configurations in $C$ whose weight is greater that zero: $\forall c \in C\ (c.w > 0 \rightarrow \exists p \in CA $ such that $p$ covers $c.pr$).
    \end{definition}

    \begin{definition}{\textbf{Coverage}.}
        Given a covering array $CA$ and a set of configurations $C$, we define $cov(CA)$ as the sum of all configuration weights in $C$ covered by any configuration in $CA$ divided by the sum of all configuration weights in $C$, that is:
        
        $$cov(\mathit{CA}) = \frac{\sum\limits_{\substack{c\in C,\ \exists p \in \mathit{CA},\ p\ covers\ c.pr}} c.w}{\sum\limits_{c\in C} c.w}$$
    \end{definition}

    The optimization problem of our interest consists in finding a covering array $CA$ with the minimum number of products $|CA|$ maximizing the coverage, $cov(CA)$.

\subsection{Construct, Merge, Solve \& Adapt}
    Matheuristics are techniques that combine metaheuristics and mathematical programming techniques. The \textit{Construct, Merge, Solve \& Adapt} (CMSA) algorithm is matheuristic for combinatorial optimization introduced by C. Blum et al. \cite{Blum:2016}. Before describing the algorithm we present some concepts.

    Given a problem $P$, let $C$ be the set of all possible components of the solution to the instance of our problem $I$. $C$ is called the complete set of solution components with respect to $I$. 
    A valid solution $S$ to $I$ is represented as a subset of the solution components, that is, $S \subseteq C$.
    Finally, a sub-instance $C'$ of $I$ is a subset of the set of solution components, so $C' \subseteq C$.

    The idea of the algorithm is the following (see Algorithm~\ref{alg:cmsa} and flow graph in Figure~\ref{fig:graph}). While the time limit established is not reached:
    \begin{enumerate}
        \item First, it generates $n_a$ solutions of the main instance of the problem $I$.
        \item All the components belonging to the generated solutions are merged to form a sub-instance of the problem, $C'$.
        \item An exact algorithm is applied to the sub-instance $C'$, expecting that the solution for $C'$ is (quasi-)optimal for $C$.
        \item The solution is compared with the best solution found at the moment and $C'$ is updated accordingly with a defined aging policy, mostly deleting useless solution components.
    \end{enumerate}
    The algorithm has two key components that have to be defined accordingly for the target problem:
    \begin{itemize}
        \item A solution generator: based on some randomized strategy and providing high quality solutions.
        \item An exact solver for the sub-instances: for example, based on ILP or other exact algorithm.
    \end{itemize}

    \begin{algorithm}
      \caption{Construct, Merge, Solve \& Adapt (CSMA)}
      \label{alg:cmsa}
      \textbf{input:} set of problem components $C$, values for parameters $n_a$ and $age_{max}$\\
      $bestSolution$ := $C$\\
      $subInstance$ := $\varnothing$\\
      age[$c$] := 0 for all $c \in C$\\
      \While{CPU time limit not reached} {
        \For{$t$ = 1, ..., $n_a$} {
            $probSolution$ := ProbabilisticSolution($C$)\\
            \For{all $x \in probSolution$ \textbf{and} $x \notin subInstance$} {
                age[$x$] := 0\\
                $subInstance$ := $subInstance \cup \{x\}$
            }
        }
        $exactSolution := $ ExactSolver($subInstance$)\\
        \If{$exactSolution$ is better than $bestSolution$} {
          $bestSolution$ := $exactSolution$\\
        }
        Adapt($subInstance$, $exactSolution$, $age_{max}$)
      }
      \textbf{output:} $bestSolution$
    \end{algorithm}

    \begin{figure}[h!]

        \centering
        \includegraphics[width=\textwidth]{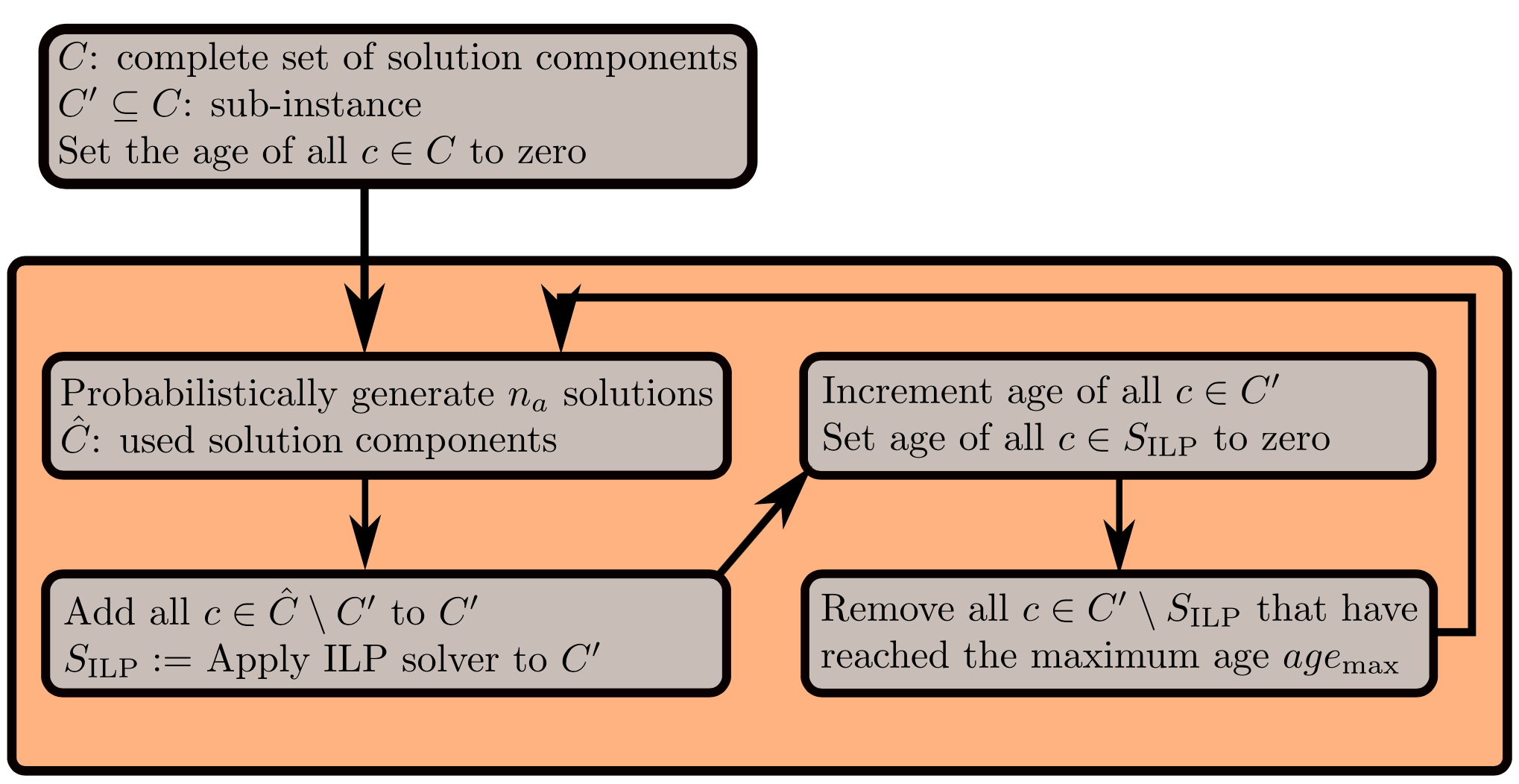}
        \caption[CMSA flow graph]{CMSA flow graph (image from C. Blum)}
    \label{fig:graph}        
    \end{figure}

%% file: Chapters/DesignAndImplementation.tex
\label{design}

In order to apply the CMSA algorithm to our problem we have to define the three methods described in algorithm \ref{alg:cmsa}: The \textit{ProbabilisticSolution}, \textit{ExactSolver} and \textit{Adapt} methods. Before that, we have to guess what is a solution component for the prioritized pairwise test data generation in the CMSA algorithm. In this case, given that we want to minimize the cardinality of the set of products that entirely covers all the possible configurations in the feature model, a solution component is a valid product from the feature model $FM$, being $P$ the set of all valid products of $FM$.

Coming up next we introduce the idea of the three components of the CMSA adaptation to our problem. First, we explain the \textit{ProbabilisticSolution} method, later on we move on to the \textit{ExactSolver} method. Finally, we describe the aging policy.

\subsection{Solution Generation}
 At the start of the main loop of the algorithm several solutions for $P$ are generated to be merged in a final sub-instance that is solved later.
We consider the whole search space $P$, generating random products that are valid within the feature model. This strategy guarantees that, eventually, the algorithm will find a solution better  than the previous one. The pseudo-code of this method is described below in algorithm \ref{alg:cmsar}.

  \begin{algorithm}[h!]
    \caption{ProbabilisticSolution}
    \label{alg:cmsar}
  
    \textbf{input:} problem instance $I$\\
    $solution$  := $\varnothing$\\
    $uncovered$ := $validPairs(I)$\\
    \While{uncovered $\neq \varnothing$} {
      $x$ := generateRandomValidProduct\\
      \If{$configurations(x) \cap uncovered \neq \varnothing$} {
        $solution$ := $solution \cup \{x\}$\\
        $uncovered$ := $uncovered - configurations(x)$\\
      }
    }
    \textbf{output:} $solution$
    
  \end{algorithm}
  
The fact that there is no uncovered configuration in each solution generated ensures that the solution will always cover all the weighted configurations, and this property also holds in the sub-instance generated after merging all the solutions.

  \subsection{Exact Solver}
  
We use an exact algorithm to compute the best solution for the sub-instance generated. 
We use an ILP model to select a subset of products for the sub-instance which, covering all weighed pairs, has minimum cardinality. This problem is equivalent to the test suite minimization problem, in its mono-objective version~\cite{Arito2012} and, thus, we can use the same ILP model to solve the problem.
  
  \subsection{Adapting the sub-instance}
  
 The aging mechanism ued here is the same proposed by Blum et al.~\cite{Blum:2016}. In each iteration of the algorithm, the sub-instance is composed by a subset $C' \subseteq C$ of the components of the problem. The \textit{ExactSolver} method returns a solution, which is a subset $S \subseteq C'$ of components.
Then, the ``age'' of all the components that are part of the solution $S$ are reset to 0, while the rest of the components see their age increased by 1. If any component reaches the maximum age established ($age_{max}$), then the component is removed from $C'$. Algorithm \ref{alg:adapt} shows the pseudo-code for the \textit{Adapt} method.

    \begin{algorithm}
      \caption{Adapt}
      \label{alg:adapt}
      \textbf{input:} sub-instance $C'$, sub-instance solution $S$\\
      \For{$x$ in $S$} {
        $age[x]$ := $0$\\
      }
      \For{$x$ in $C' - S$} {
        $age[x]$ := $age[x] + 1$\\
        \If{$age[x]$ equals $age_{max}$} {
          $C'$ := $C' - \{x\}$\\
        }
      }
      \textbf{output:} $C'$
    \end{algorithm}


%% file: Chapters/Results.tex
\label{experiments}

In this section we describe how the analysis of the approach is performed. First, we introduce the other four algorithms we compare with CMSA. Later on, we describe the benchmark used for the evaluation. Finally, we explain the different experiment configurations.

\subsection{Hybrid Algorithms Based on Integer Programming}

Two different hybrid algorithms combining a greedy heuristic and integer programming were introduced by Ferrer et al.~\cite{Ferrer2017}. The first one, called HILP, is based on an integer linear formulation, and the second, named HINLP, is based on a quadratic (nonlinear) integer formulation. The two algorithms proposed in this work use the same high level greedy strategy. In each iteration they try to find a product that maximizes the weighted coverage. They select in each iteration the product that contributes with greater coverage to the actual solution. The algorithm applies the heuristic to the whole product set (that can be of billions of possible products) instead of small subsets. For further details on HILP or HINLP, please refer to~\cite{Ferrer2017}. 

\subsection{Prioritized Pairwise Genetic Solver}
\label{subsec:geneticsolver}

\emph{Prioritized Pairwise Genetic Solver} (PPGS) is a constructive genetic algorithm that follows a master-slave model to parallelize the individuals' evaluation. In each iteration, the algorithm adds the best product to the test suite until all weighted pairs are covered. The best product to be added is the product that adds more weighted coverage (only pairs not covered yet) to the set of products. 

The parameter setting used by PPGS is the same of the reference paper for the algorithm~\cite{Lopez-Herrejon:2014:PEA:2576768.2598305}. It uses binary tournament selection and a one-point crossover with a probability $0.8$. The population size of 10 individuals favours the exploitation rather than the exploration during search. The termination condition is to reach 1,000 fitness evaluations. The mutation operator iterates over all selected features of an individual and randomly replaces a feature by another one with a probability $0.1$. The algorithm stops when all the weighted pairs have been covered. For further details on PPGS see~\cite{Lopez-Herrejon:2014:PEA:2576768.2598305}.

\subsection{Prioritized-ICPL Algorithm}
\label{subsec:icpl}

Prioritized-ICPL (pICPL) is a greedy algorithm to generate \emph{n}-wise covering arrays proposed by Johansen et al.~\cite{DBLP:conf/models/JohansenHFES12}. pICPL does not compute covering arrays with full coverage but rather covers only those \emph{n}-wise combinations among features that are present in at least one of the prioritized products, as was described in the formalization of the problem in Section~\ref{sec:problem}.
We must highlight here that the pICPL algorithm uses \emph{data parallel execution}, supporting any number of processors. Their parallelism comes from simultaneous operations across large sets of data. For further details on prioritized-ICPL please refer to~\cite{DBLP:conf/models/JohansenHFES12}.

\subsection{Benchmark}

The feature models that we use for the comparison of the algorithms are generated from 16 real SPL systems. 
We considered a method called \textit{measured values} to assign weight values to prioritized products. This method consists in assigning the weights derived from non-functional property values obtained from 16 real SPL systems, that were measured with the SPL Conqueror approach introduced by Siegmund et al.~\cite{SIEGMUND2013491}. This approach aims at providing reliable estimates of measurable non-functional properties such as performance, main memory consumption, and footprint. These estimations are then used to emulate more realistic scenarios where software testers need to schedule their testing effort giving priority, for instance, to products or feature combinations that exhibit higher footprint or performance. In this work, we use the actual values taken on the measured products considering pairwise feature interactions. Table~\ref{tab:dataset} summarizes the SPL  systems evaluated, their feature number (\texttt{FN}), products number (\texttt{PN}), configurations number measured (\texttt{CN}), and the percentage of prioritized products (\texttt{PP$\%$}) used in our comparison.

 \begin{center}
    \captionof{table}{Benchmark of feature models}
    \label{tab:dataset}
    \begin{tabular}{| l | r | r | r | r |}
        \hline
         \textbf{Model name}   & \textbf{FN} &    \textbf{PN} & \textbf{CN} & \textbf{PP$\%$}\\\hline
         Apache                &       $ 10$ & $         256$ &      $ 192$ &    $      75.0$\\\hline
         BerkeleyDBFootprint   &       $  9$ & $         256$ &      $ 256$ &    $     100.0$\\\hline
         BerkeleyDBMemory      &       $ 19$ & $        3840$ &      $1280$ &    $      33.3$\\\hline
         BerkeleyDBPerformance &       $ 27$ & $        1440$ &      $ 180$ &    $     12.50$\\\hline
         Curl                  &       $ 14$ & $        1024$ &      $  68$ &    $       6.6$\\\hline
         LinkedList            &       $ 26$ & $        1440$ &      $ 204$ &    $      14.1$\\\hline
         Linux                 &       $ 25$ & $  \approx3E7$ &      $ 100$ &    $\approx0.0$\\\hline
         LLVM                  &       $ 12$ & $        1024$ &      $  53$ &    $       5.1$\\\hline
         PKJab                 &       $ 12$ & $          72$ &      $  72$ &    $     100.0$\\\hline
         Prevayler             &       $  6$ & $          32$ &      $  24$ &    $      75.0$\\\hline
         SensorNetwork         &       $ 27$ & $       16704$ &      $3240$ &    $      19.4$\\\hline
         SQLiteMemory          &       $ 40$ & $  \approx5E7$ &      $ 418$ &    $\approx0.0$\\\hline
         Violet                &       $101$ & $ \approx1E20$ &      $ 101$ &    $\approx0.0$\\\hline
         Wget                  &       $ 17$ & $        8192$ &      $  94$ &    $      1.15$\\\hline
         x264                  &       $ 17$ & $        2048$ &      $  77$ &    $       3.7$\\\hline
         ZipMe                 &       $  8$ & $          64$ &      $  64$ &    $     100.0$\\\hline
    \end{tabular}
    \end{center}
    
In the case of the feature models where the percentage of the prioritized products is equal to 100$\%$, applying the heuristic to the whole solution components set $P$ (without generating sub-instances) should give similar results than applying HINLP. A greedy implementation of CMSA without generating sub-instances have been implemented in order to test the rest of the functionalities of the algorithm and validate the consistency of the results.

\subsection{Experiments Configuration}

The experiments were run on a cluster of 16 machines with Intel Core2 Quad
processors Q9400 (4 cores per processor) at 2.66$\,$GHz and 4$\,$GB memory and
2 nodes (96 cores) equipped with two Intel Xeon CPU (E5-2670 v3) at 2.30$\,$GHz
and 64$\,$GB memory. The cluster was managed by HTCondor 8.2.7, which allowed
us to perform parallel independent executions to reduce the overall experimentation
time. 

For the CMSA algorithm, different parameter configurations of time limit ($max_{time}$), solutions generated per iteration ($n_a$) and maximum age for the aging policy ($age_{max}$) were used. As a result, we chose the best configuration with $n_a$=5 and $age_{max}=4$. For each configuration, a total of 30 independent runs were executed.

We have to keep in mind that the $max_{time}=$3 seconds indicates the limit of time for the last iteration, in cases of some feature models (i.e. Violet) where the number of constraints is high, the algorithm can take more time. We applied the Kolmogorov-Smirnov normality test to confirm that the distribution of the results is not normal. Therefore, we applied the non-parametric Kruskal-Wallis test with a confidence level of 95\% (\emph{p}-value under $0.05$) with Bonferroni's 
$p$-value correction to check if the observed differences are statistically significant. In the cases where Kruskal-Wallis test rejects the null hypothesis, we run a single factor ANOVA post hoc test for pairwise comparisons.

\section{Analysis of the results}
\label{results}

In this section we analyze the results of the execution of CMSA in comparison to the results of state-of-the-art algorithms (HILP, HINLP, PPGS and pICPL). Table \ref{tab:coverage} summarizes the results of the execution of the algorithms for different values of weighted coverage. Each column corresponds to one algorithm and in the rows we show the number of products required to reach 50\% up to 100\% of weighted coverage. The data shown in each cell is the mean and the standard deviation of the number of products required to reach the coverage of the line for all the independent runs of the benchmark of feature models. We also show the average and standard deviation of the required runtime in the last line. We highlight the best value for each percentage of weighted coverage and the shortest execution time.

We observe that CMSA is the best in solution quality (less products required to obtain a particular level of coverage) for all percentages of weighted coverage. After CMSA, the algorithms based on integer programming obtain the best solutions. The difference in quality between HILP and HINLP are almost insignificant, except for 100\% coverage, so it is difficult to claim that one algorithm is better than the other. Then, PPGS is the fourth algorithm in our comparison, and finally pICPL is the worst.

\begin{table}[h!]
\centering
\caption{Mean and standard deviation of number of products and time in 30 independent runs. The best result for each percentage of prioritized coverage and total time are highlighted.}
\label{tab:coverage}
\begin{tabular}{llllll}
\hline
Coverage & \multicolumn{1}{c}{CMSA} & \multicolumn{1}{c}{HILP} & \multicolumn{1}{c}{HINLP} & \multicolumn{1}{c}{PPGS} & \multicolumn{1}{c}{pICPL} \\ \hline
50\% & \cellcolor{gray95}$1.56_{0.50}$ & \cellcolor{gray95}$1.56_{0.50}$ & \cellcolor{gray95}$1.56_{0.50}    $ & $1.58_{0.49}$ & \cellcolor{gray95} $1.56_{0.50} $ \\ \hline
75\% & \cellcolor{gray95}$2.53_{0.71}$ & $2.63_{0.78} $ & $2.63_{0.78}$ & $2.66_{0.77} $ & $2.75_{0.75} $ \\
80\% & \cellcolor{gray95}$2.75_{0.75}$ & $2.81_{0.81} $ & $2.81_{0.81}$ & $2.81_{0.73}$ & $3.25_{0.97} $ \\
85\% & \cellcolor{gray95}$3.31_{0.87}$ & $3.44_{0.86}$ & $3.44_{0.86}$ & $3.46_{0.87}$ & $3.81_{0.95} $ \\ \hline
90\% & \cellcolor{gray95}$3.79_{0.77}$ & $4.06_{1.03}   $ & $4.00_{0.94}$ & $4.12_{1.04} $ & $4.56_{1.27}$ \\
95\% & \cellcolor{gray95}$4.94_{0.90}$ & $5.37_{1.05}   $ & $5.38_{1.05} $ & $5.45_{1.14} $ & $6.06_{1.44} $ \\
96\% & \cellcolor{gray95}$5.17_{1.05}$ & $5.69_{1.16}   $ & $5.69_{1.16}  $ & $5.86_{1.18} $ & $6.38_{1.58} $ \\
97\% & \cellcolor{gray95}$5.67_{1.09}$ & $6.13_{1.22}   $ & $6.13_{1.22}  $ & $6.24_{1.38} $ & $6.75_{1.39} $ \\
98\% & \cellcolor{gray95}$5.96_{1.26}$ & $6.81_{1.42}   $ & $6.75_{1.39} $ & $6.98_{1.55} $ & $7.44_{1.66} $ \\
99\% & \cellcolor{gray95}$6.79_{1.52}$ & $7.75_{1.64}   $ & $7.75_{1.64} $ & $7.92_{1.87} $ & $8.75_{2.08} $ \\
100\%& \cellcolor{gray95}$10.06_{4.99}$ & $11.69_{5.51}  $ & $11.63_{5.33} $ & $12.08_{6.50} $ & $12.19_{5.68} $ \\ \hline
Time (s) & $164_{346}$ & $18_{10}$ & \cellcolor{gray95}$1_{2}$ & $2049_{7684}$ & $24_{59}$ \\ \hline
\end{tabular}
\end{table}

Regarding the execution time, we can appreciate that HINLP is clearly the fastest algorithm, actually thanks to the nonlinear formulation it produces a boost in computation time due to the reduction of clauses in comparison with the linear variant of the algorithm. It is closely followed  by HILP and pICPL (also based on a greedy strategy). They are followed by CMSA in the speed ranking and, finally, quite far from the rest, PPGS, a genetic algorithm. 

In order to check whether the differences between the algorithms are statistically significant or just a matter of chance, we have applied the statistical tests explained in the previous section. For 50\% coverage there is no significant differences between CMSA and the others. Next, for 75\% up to 85\% of weighted coverage, there are significant differences between CMSA and pICPL. Finally, for 90\% up to 100\% CMSA is statistically significantly better than the other algorithms. Regarding the execution time, HINLP is statistically better than the other algorithms.

In Table~\ref{tab:number} we show the number of times an algorithm obtains the minimum mean value for each percentage of coverage and for the 16 realistic feature models. Note that Table~\ref{tab:number} summarizes the results showed in Table~\ref{tab:big} in the Appendix. On the one hand, we observe that in only 3 out of 176 comparisons (16 feature models and 11 percentages of weighted coverage) other algorithm different than CMSA has obtained a better value for solution quality. On the other hand, in 173 out of 176 comparisons CMSA obtains the best results of the comparison, moreover in 70 out 176 CMSA is the only obtaining the best value, i.e. , no other algorithm obtains such a low value. In general, it can also be seen in this table that, the larger the value of weighted coverage, the better the results of CMSA. The reason behind this behavior is that in early stages of the search, for small and medium values of coverage, it is easier to find the products which add not-covered-yet pairs. When the search progresses, it is harder to find the products which are able to add not-covered-yet pairs.

\begin{table}[h!]
\centering
\caption{Times an algorithm has the best average number of products per percentage of coverage.}
\label{tab:number}
\begin{tabular}{l|ccccc}
\hline
\multicolumn{1}{c|}{Coverage} & CMSA & HILP & HINLP & PPGS & pICPL \\ \hline
50\% & 16 & 16 & 16 & 14 & 16 \\
75\% & 16 & 13 & 13 & 10 & 11 \\
80\% & 16 & 15 & 15 & 14 & 9 \\
85\% & 16 & 14 & 14 & 11 & 10 \\
90\% & 16 & 12 & 12 & 8 & 7 \\
95\% & 15 & 8 & 8 & 4 & 3 \\
96\% & 16 & 6 & 6 & 3 & 3 \\
97\% & 16 & 8 & 8 & 4 & 3 \\
98\% & 15 & 2 & 2 & 0 & 1 \\
99\% & 16 & 2 & 2 & 1 & 2 \\
100\% & 15 & 2 & 2 & 1 & 3 \\ \hline
Total & 173 & 98 & 98 & 70 & 68 \\ \hline
\end{tabular}
\end{table}

Let us now focus on how the algorithms obtain total weighted coverage in each feature model. In Table~\ref{tab:instances} we show the mean value for 100\% weighted coverage. We also show the standard deviation in the cases it is not zero (when the algorithm is randomized and different values are obtained in at least two of the 30 independent runs). In most feature models (15 out of 16) CMSA obtains the best value. The exception is in the Linux feature model, where pICPL is the best. In three models (Curl, Prevayler and Violet) CMSA obtains the best value, although at least other algorithm is able to reach the same value. In the rest of the feature models analyzed here (12 out of 16) no other algorithm reach the same solution quality than CMSA. As we expected, there are significant differences between CMSA and the other proposals in the pairwise comparisons. Specifically, in 52 out of 64 comparisons (16 feature models and 4 algorithms) CMSA is significantly better than the other algorithms, in 11 out of 64 there is no difference between CMSA and other algorithm, and only once it is significantly worse, particularly in the Linux model with pICPL.

\begin{table}[]
\centering
\caption{Mean values over 30 independent runs for CMSA, HILP, HINLP, PPGS and pICPL. The best value for each FM is highlighted.}
\label{tab:instances}
\begin{tabular}{l|rrrrr}
\hline
Feature Model & \multicolumn{1}{c}{CMSA} & \multicolumn{1}{c}{HILP} & \multicolumn{1}{c}{HINLP} & \multicolumn{1}{c}{PPGS} & \multicolumn{1}{c}{pICPL} \\ \hline
Apache & \cellcolor{gray95}6.00 & 7.00 & 7.00 & 7.00 & 8.00 \\
BerkeleyDBFootprint & \cellcolor{gray95}$6.07_{0.25}$ & 8.00 & 8.00 & $8.17_{0.38}$ & 9.00 \\
BerkeleyDBMemory & \cellcolor{gray95}$20.03_{0.18}$ & 21.00 & 21.00 & $23.33_{1.06}$ & 21.00 \\
BerkeleyDBPerformance & \cellcolor{gray95}9.00 & 10.00 & 10.00 & $10.60_{0.50}$ & 12.00 \\
Curl & \cellcolor{gray95}8.00 & 9.00 & 9.00 & $9.63_{0.67}$ & \cellcolor{gray95}8.00 \\
LinkedList & \cellcolor{gray95}$11.10_{0.30}$ & 13.00 & 13.00 & $13.37_{0.49}$ & 14.00 \\
Linux & 11.00 & 11.00 & 11.00 & $11.10_{0.66}$ & \cellcolor{gray95}10.00 \\
LLVM & \cellcolor{gray95}$6.90_{0.30}$ & 10.00 & 10.00 & $8.17_{0.46}$ & 8.00 \\
PKJab & \cellcolor{gray95}6.00 & 7.00 & 7.00 & 7.00 & 8.00 \\
Prevayler & \cellcolor{gray95}6.00 & \cellcolor{gray95}6.00 & \cellcolor{gray95}6.00 & \cellcolor{gray95}6.00 & \cellcolor{gray95}6.00 \\
SensorNetwork & \cellcolor{gray95}10.00 & 14.00 & 14.00 & $13.97_{1.16}$ & 17.00 \\
SQLiteMemory & \cellcolor{gray95}$23.87_{1.41}$ & 28.00 & 27.00 & $31.53_{1.99}$ & 28.00 \\
Violet & \cellcolor{gray95}12.00 & \cellcolor{gray95}12.00 & \cellcolor{gray95}12.00 & $12.83_{0.59}$ & 15.00 \\
Wget & \cellcolor{gray95}$9.87_{0.35}$ & 12.00 & 12.00 & $11.37_{1.00}$ & 11.00 \\
x264 & \cellcolor{gray95}$9.07_{0.25}$ & 12.00 & 12.00 & $12.10_{1.03}$ & 13.00 \\
ZipMe & \cellcolor{gray95}6.00 & 7.00 & 7.00 & $7.03_{0.18}$ & 7.00 \\ \hline
Total & \cellcolor{gray95}$10.06_{4.99}$ & $11.69_{5.51}$ & $11.63_{5.33}$ & $12.08_{6.50}$ & $12.19_{5.68}$ \\ \hline
\end{tabular}
\end{table}

In the comparison between CMSA and HINLP (the second best algorithm), we can appreciate that HINLP is doubtless faster, by contrast, in solution quality it is only able to find the best known solution in two feature models. In addition, there are significant differences between CMSA and HINLP in 14 out of 16 models, so we can claim that CMSA is definitely the best algorithm in the comparison in regards to solution quality.

\begin{figure}[h]
        \centering
        \includegraphics[width=8cm]{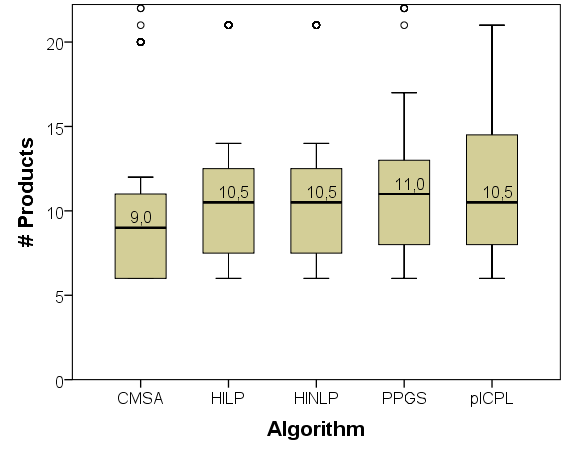}
        \caption{Number of products needed to achieve total coverage. For each feature model (16), there are 30 solutions obtained by independent replications.}
        \label{fig:boxplot}
    \end{figure}

In order to illustrate the comparison regarding the solution quality for total weighted coverage, in Figure~\ref{fig:boxplot} we show a boxplot for each algorithm considering all the feature models. Note that several outliers for all algorithms are outside the worst range shown in the plot. The boxplot confirms again that CMSA is the best for 100\% weighted coverage with a median value of 9 products, followed by HILP and HINLP with a median of 10,5 products, and PPGS with a median of 11 products. In addition, we can also appreciate that CMSA has a lower interquartile range, and all the quartile marks are lower in comparison to the other algorithms. 

%% file: Chapters/Conclusions.tex
\label{conclusiones}
\label{conclusions}

In this work we have applied a novel matheuristic approach (CMSA) to the Prioritized Pairwise Test Data Generation Problem, aiming to ease the task of testing on large SPLs. Our main contribution is the adaptation of the CMSA algorithm to this problem for SPL, relating the CMSA algorithm to the specific nuances of the problem.

We present the empirical results derived from the evaluation of our CMSA approach on the introduced benchmark of feature models. We compare CMSA with four different approaches to tackle the problem of prioritized pairwise test data generation for SPL. Regarding the solution quality, our analysis showed an improvement in terms of the quality metric, which is better —i.e., lower in terms of the number of products— in almost all instances of the benchmark and for all percentages of weighted coverage. In addition, in most comparisons the test suites computed by CMSA are statistically significantly better than those computed by the other algorithms.

Testing on a SPL means a high cost in resources and time due to the effort devoted to the testing phase of even one single product, which can require several hours. Therefore, it is straightforward to think that the best approach is the one that reduces the size of the test suite, in this case our proposal: CMSA. In addition, the execution of the algorithm only requires a few minutes, much less than testing one single product in most of the scenarios. A general conclusion is that CMSA is clearly the best approach for computing prioritized pairwise test data. On the other hand, the approach based on nonlinear integer programming (HINLP) is able to obtain good quality solutions in only a few seconds, then it is the best option when a good solution is immediately needed. This could be the case when testing a single product of the SPL only requires a few seconds.

There remains one Achilles' heel clearly identified in our proposal, the execution time is higher than several approaches studied here. Future work will require a deeper analysis of the performance and quality of the generation of random products, that is the baseline of the construct phase of CMSA. We plan to assess whether CPLEX, the optimizer used to solve the integer programming problems found in this work, is able to provide enough different products to obtain the diversity that every search algorithm requires.